# Adaptive Explainable Neural Networks (AxNNs)

Jie Chen[1], Joel Vaughan, Vijayan N. Nair, and Agus Sudjianto[2]

**Corporate Model Risk, Wells Fargo, USA**

Jun 1, 2020

## Abstract

While machine learning techniques have been successfully applied in several fields, the black-box nature of the models presents challenges for interpreting and explaining the results. We develop a new framework called Adaptive Explainable Neural Networks (AxNN) for achieving the dual goals of good predictive performance and model interpretability. For predictive performance, we build a structured neural network made up of ensembles of generalized additive model networks and additive index models (through explainable neural networks) using a two-stage process. This can be done using either a boosting or a stacking ensemble. For interpretability, we show how to decompose the results of AxNN into main effects and higher-order interaction effects. The computations are inherited from Google's open source tool AdaNet and can be efficiently accelerated by training with distributed computing. The results are illustrated on simulated and real datasets.

Keywords: Additive index models, Boosting, Generalized additive models, Interpretable machine learning, Main effects and Interactions, Stacking

Abbreviations used in the paper:
- AIM: Additive Index Model
- AxNN: Adaptive explainable neural network
- FFNN: Feedforward neural networks
- GAM: Generalized additive model
- GAMnet: Generalized additive model network
- ML: Machine learning
- NN: Neural network
- RF: Random forest
- XGB: XgBoost
- xNN: Explainable Neural Network

---

[1] email: Jie.Chen@wellsfargo.com

# 1   Introduction

Machine learning (ML) algorithms provide significant advantages of over traditional data analysis methods for several reasons: i) availability of scalable algorithms to handle large datasets; ii) ability to automate much of the routine data analysis and modeling tasks; and iii) develop flexible models with excellent predictive power. However, the techniques are complex and the results are not easy to understand, interpret, and explain. This poses a serious problem in a number of application areas, especially in regulated industries like banking.

There has been extensive research to develop diagnostics that can be used to do post-hoc interpretation and explanation of complex ML algorithms. Early work on diagnostic tools for studying input-output relationships include partial dependence plots (J. H. Friedman 2001) and individual conditional expectation plots by (Goldstein, et al. 2015). More recent work includes accumulated local effect plots (Apley 2016) and accumulated total derivative effect plots (Liu, et al. 2018).

There are also attempts to explicitly build in explainability into the architecture of ML algorithms. GA2M, a fast algorithm based on special tree structures, was introduced in (Lou, et al. 2013, Caruana, et al. 2015) to include pairwise interactions and thereby extend generalized additive models (GAMs). (Vaughan, et al. 2018, Yang, Zhang and Sudjianto 2019) developed a structured neural network (called explainable neural network or xNN) based on additive index models or AIMs. (Tsang, Liu, et al. 2018) proposed neural interaction transparency to disentangle shared-learning across different interactions via a special neural network structure.

This paper proposes a new approach aimed at simultaneously achieving good predictive performance and model interpretability. The technique is based on a two-stage framework that called `adaptive explainable neural network' (AxNN). In the first stage, an ensemble with base learners of generalized additive model networks (GAMnet) is used to capture the main effects. In the second stage, an ensemble of explainable neural networks (xNN), that are incremental to the first stage, is used to adaptively fit additive index models (AIMs). The differences between the two stages can be interpreted as interaction effects, allowing for direct interpretation of the fitted model. Our flexible implementation allows model-fitting through boosting or stacking. Both of them have similar predictive performance.

AxNN does not require extensive tuning of the neural networks. This is a major advantage as hyper-parameter optimization can be computationally intensive and has been the subject of considerable focus in the ML literature. Many AutoML approaches have been developed to avoid manual tuning by automatic learning and optimization. Examples include auto-sklearn (Feurer, et al. 2015), auto-pytorch (Mendoza, et al. 2016), and AdaNet (Weill, et al. 2019). Among them, AdaNet is especially relevant for our work. It is an adaptive algorithm for learning a neural architecture. Our computational engine for AxNN is built using Google's AdaNet implementation, and it inherits the benefits of efficient neural network architecture search by adaptively learning via multiple subnetwork candidates. See Section 2.2 for more discussion.

The remaining part of this paper is organized as follows. Section 2 gives an overview of related work. Section 3 discusses our AxNN formulation, describes the model-fitting algorithms for stacking and boosting ensembles as well as the ridge decomposition to identify main and interaction effects. Sections 4 and 5 demonstrate the usefulness of the results on synthetic and real datasets. The paper concludes with remarks in Section 6.

## 2 Review of related work

### 2.1 Explainable neural network (xNN)

Explainable neural network (xNN), proposed in (Vaughan, et al. 2018), is based on the additive multiple index model (AIM):

$$f(\mathbf{x}) = g_1(\boldsymbol{\beta}_1^T \mathbf{x}) + g_2(\boldsymbol{\beta}_2^T \mathbf{x}) + \ldots + g_K(\boldsymbol{\beta}_K^T \mathbf{x}) \qquad (1)$$

where $g_k(.), k = 1, \ldots, K$ are often referred as ridge functions, $\beta_k$ as projection indices, $\mathbf{x}$ is a $P$-dimensional covariate. The $K = 1$ case is called a single index model (Dudeja and Hsu 2018). There are many ways of estimating AIMs, and the earliest methodology was called projection pursuit (Friedman and Stuetzle 1981). See (Ruan and Yuan 2010) for penalized least-squares estimation and (Yang, et al. 2017) for the use of Stein's method.

The xNN approach in (Vaughan, et al. 2018) uses a structured neural network (NN) and provides a direct approach for model-fitting via gradient-based training methods. Figure 1 illustrates the architecture with three structural components: (i) the projection layer (first hidden layer) $\boldsymbol{\beta}_k^T x$ with linear activation functions (ii) subnetwork $g_k(\cdot)$ that is a fully connected, multi-layer neural network with nonlinear activation functions (e.g., RELU), and (iii) the combination layer $f(x)$ that computes a weighted sum of the output of ridge functions.

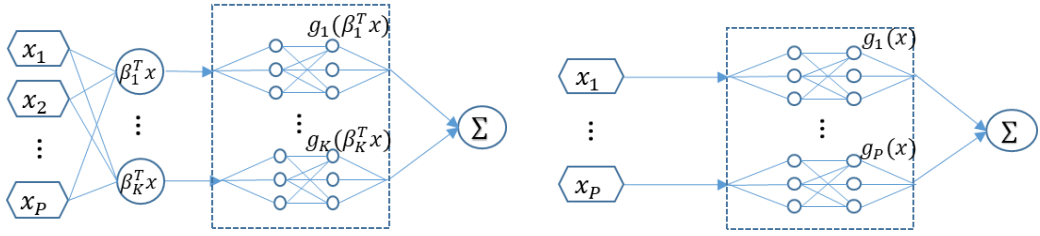

Figure 1: Illustration of structured NN architectures: xNN (left) and GAMnet (right)

The xNN-based formulation in (Vaughan, et al. 2018) is computationally fast since it uses available efficient algorithms for NNs. It is trained using the same mini–batch gradient–based methods and is easy to fit on large datasets. Further, one can take advantage of advances in modern computing such as GPUs. However, it requires careful hyper-parameter tuning which is often computationally intensive.

Generalized additive model network (GAMnet) are special cases of xNN used to estimate the following GAM structure using NNs:

$$f(\mathbf{x}) = g_1(x_1) + g_2(x_2) + \ldots + g_P(x_P). \qquad (2)$$

In this case, each ridge function in Figure 1 has only a one-dimensional input and captures just the main effect of the corresponding predictor.

### 2.2 AdaNet

As noted earlier, the computational foundation of our AxNN approach relies on the AdaNet algorithm (Cortes, Gonzalvo, et al. 2017). AdaNet aims to directly minimize the DeepBoost generalization bound (Cortes, Mohri and Syed 2014) when applied to neural networks. It does this by growing an NN as an ensemble of subnetworks that minimizes the objective

$$F(w) = \frac{1}{N}\sum_{i=1}^{N} \Phi(\sum_{j=1}^{J} w_j\, h_j(\boldsymbol{x_i}), y_i) + \sum_{j=1}^{J}(\lambda r(h_j) + \beta)|w_j|, \qquad (3)$$

Where $N$ is number of sample size, $\Phi$ is the loss function, $J$ is the number of iterations, $w_j$ is the weight for each base learner, $h_j$ is the base learner, $\boldsymbol{x_i} = (x_{i,1}, \ldots, x_{i,P})^T$ is the set of predictors, $r(h_j)$ is a complexity measure based on Rademacher complexity approximation, $\lambda$ and $\beta$ are tunable hyperparameters, and the second summation in (3) is the regularization term for network complexity.

AdaNet grows the ensemble of NNs adaptively. At each iteration, it measures the ensemble loss for multiple candidates and selects the best one to move onto for the next iteration. For subsequent iterations, the structure of the previous subnetworks is frozen, and only newly added subnetworks are trained. AdaNet is capable of adding subnetworks of different depths and widths to create a diverse ensemble, and trades off performance improvement with the number of parameters.

Current applications of AdaNet are based on feedforward NNs (FFFNs), convolution or recurrent NNs. The results from the latter two can be difficult to interpret, so we restrict attention to FFNNs. Our AxNN inherits its performance and tuning advantages from AdaNet. So, AxNN is computationally efficient and scalable, and the computations can be easily accelerated with distributed CPU, GPU, or even TPU hardware.

## 3 Adaptive explainable neural networks (AxNNs)

### 3.1 Formulation

There are many approaches in the literature for formulating and fitting a statistical model in terms of interpretable components. In particular, the notions of main effects and higher-order interactions have been around since at least the 1920s when two-way ANOVA was introduced. These concepts have been extended over the years to more complex models with many different definitions (see for example, Friedman & Popescu, 2008; Sorokina, et al. 2008; Dodge and Commenges 2006; and Tsang, Cheng and Liu 2017). There are also a variety of approaches to estimating these components (Sobol 1993; Tibshirani 1996; Hooker 2004; Hooker 2007; Friedman & Popescu,2008; Bien 2013; and Purushotham 2014).

Before going further, it is worth reiterating that it is rarely possible to reconstruct an underlying model form exactly based on data. The fitted predictive model is an approximation and the form will depend on the particular architecture used, whether it is simple parametric models, semi-parametric models such as splines, or non-parametric machine learning algorithms such as random forest, gradiaent boosting, or neural networks. Our AxNN approach is no different.

The formulation and underlying architecture of AxNN is described below. Let

$$h(E(Y|x)) = f(\boldsymbol{x}) = f(x_1, \ldots, x_P)$$

denote the overall model that captures the effects of all $P$ predictors, conditional on a fixed set of values the predictors. Here, $h(\cdot)$ is a link function such as the logit for binary responses. Further, let $g_p(x_p)$ be the main effect that represents the effect of $x_p$, averaged over all the other predictors, for $p = 1, \ldots, P$. Define the difference:

$$I(x) = f(x) - [g_1(x_1) + \cdots + g_P(x_P)]. \quad (4)$$

AxNN consists of: a) first fitting the main effects using GAMnet; and ii) then using xNN to fit an AIM to capture the remaining structure in $I(x)$ in equation (4).

The paper further shows how the fitted results can be decomposed into main effects and higher-order interaction effects. It also develops diagnostics that can be used to visualize the input-output relationships, similar to partial dependence plots (PDPs). However, unlike the PDPs which are post-hoc diagnostic tools, the main and interaction effects from AxNN are obtained directly from decomposing the ridge functions of the AxNN algorithm. One side benefit is that they do not suffer from the extrapolation concerns in the presence of correlated predictors discussed in (Liu, et al. 2018). We also provide an importance measure for ranking the significance of all the detected main and interaction effects.

Since there is some ambiguity in the literature, we note that the term main effect is used here to denote the effect associated with an individual predictor obtained by projecting the original model onto the space spanned by GAMs. To make this concrete, consider following simple model with independent predictors. Let

$$f(x) = \beta_0 + \beta_1 x_1 + \beta_2 x_2 + \beta_{12} x_1^2 x_2^2.$$

Further, let $c_i = E(X_i^2), i = 1, 2$, where the expectation is taken with respect to the distribution of the predictors. Then, fitting a GAM will estimate the main effects $(\beta_1 x_1 + \beta_{12} c_2 x_1^2)$, and $(\beta_2 x_2 + \beta_{12} c_1 x_2^2)$ respectively for the two predictors. These are quadratic while the quadratic terms appear only in the interaction in the model above. Further, the residual interaction term is $I(x) = \beta_{12}(x_1^2 - c_1)(x_2^2 - c_2)$. Note that it satisfies the usual condition that its marginal expectation with respect to the distribution of each predictor is zero. In this simple case, the original model can be reconstructed exactly by algebraic re-arrangement of the main effects and interactions.

But this will not be the case in most realistic situations. Consider, for example, the function $f(x) = \log(x_1 + x_2)$ with $a < x_i < b, i = 1, 2$. For suitable values of $(a, b)$, the function $\log(x_1 + x_2)$ can be approximated very well by a GAM. In this case, residual interaction term will be small. We demonstrate this phenomenon through more complex examples in Section 4. As noted earlier, when the underlying models are complex, it is not possible to recover the true form exactly. The fitted model will depend on the architecture. Sections 4 and 5 shows how AxNN works and that it provides excellent insights into the underlying structure.

### 3.2 Algorithms for fitting AxNN with Boosting

As described in the AdaNet formulation, our goal is to minimize the objective function

$$F(w, h) = \frac{1}{N} \sum_{i=1}^{N} \Phi\left(\sum_{j=1}^{J_1} w_j h_j(x_i) + \sum_{j=J_1+1}^{J_2} w_j h_j(x_i), y_i\right) + \sum_{j=1}^{J_2} (\lambda r(h_j) + \beta)|w_j| \quad (5)$$

where $J_1$ is the number of base learners (i.e., iterations) for first GAMnet stage, and $J_2$ is total number of base learners for both GAMnet and xNN stages. We introduce the AxNN framework using boosting or stacking ensembles and then describe the decomposition of the ridge functions into main effects and interactions.

### 3.2.1 AxNN with Boosting

We first introduce AxNN using the boosting ensemble approach. As shown in Figure 2 and described in Algorithm 1, we use GAMs in the first stage (more precisely GAMnets) as base learners to capture the main effects. Specifically, in equation (5), for the first set of ridge functions, we take

$$h_j(\boldsymbol{x}_i) = \sum_{p=1}^{P} g_{j,p}(x_{i,p}), j = 1, \ldots, J_1,$$

where $g_{j,p}(\cdot)$ is the ridge function for $x_p$ at $j$th iteration, which is modeled via a fully connected, multi-layer neural network using nonlinear activation functions (e.g., RELU). For each iteration in Stage 1, we first train the base learner $h_j(\boldsymbol{x}_i)$ by fixing the ensemble $\sum_{j=1}^{k-1} w_j h_j(\boldsymbol{x}_i)$ learned from previous boosting iterations. We then optimize the weights of from both the previous and current iterations.

When the validation performance converges, we move to Stage 2 where we use xNNs as base learners to capture the remaining effects in Equation (4). Specifically, for the second set of ridge functions in Equation (5), we use

$$h_j(\boldsymbol{x}_i) = \sum_{k=1}^{K_j} g_{j,k}(\boldsymbol{\beta}_{j,k}^T \boldsymbol{x}_i), j = J_1 + 1, \ldots, J_2$$

where $\boldsymbol{\beta}_{j,k} = (\beta_{j,k,1}, \ldots, \beta_{j,k,P})^T$. We learn these base learners incrementally: for each iteration, we first train $h_j(\boldsymbol{x}_i)$ by fixing the ensemble learned from the first stage and all previous iterations in the second stage. Then we fix all the base learners and optimize the weights. Note that we do not re-optimize the weights of GAMnet base learners from the first stage, as the estimation problem is over-parameterized: GAMnets are a subset of xNNs, so their effects can be reduced and the main effects can be absorbed by xNNs.

Following the approach in AdaNet, in each iteration, multiple networks with different architectures can be considered as candidates, and the best one will be picked up with the goal of minimizing the objective function in Equation (5).

Boosting requires the use of weak learners so that the bias can be reduced over the iterations. Therefore, the architecture of the subnetworks we use for the ridge functions in GAMnet and xNNs should be shallow and narrow (i.e., a small number of layers and number of units in the layer). Moreover, these xNNs should have a small number of ridge functions, such as a single index model (SIM) structure.

Figure 2 is an example of the flow of an AxNN architecture. The subnetworks for the ridge functions can vary with different number of layers and width. The base learners for the second stage can be single index or multiple index models.

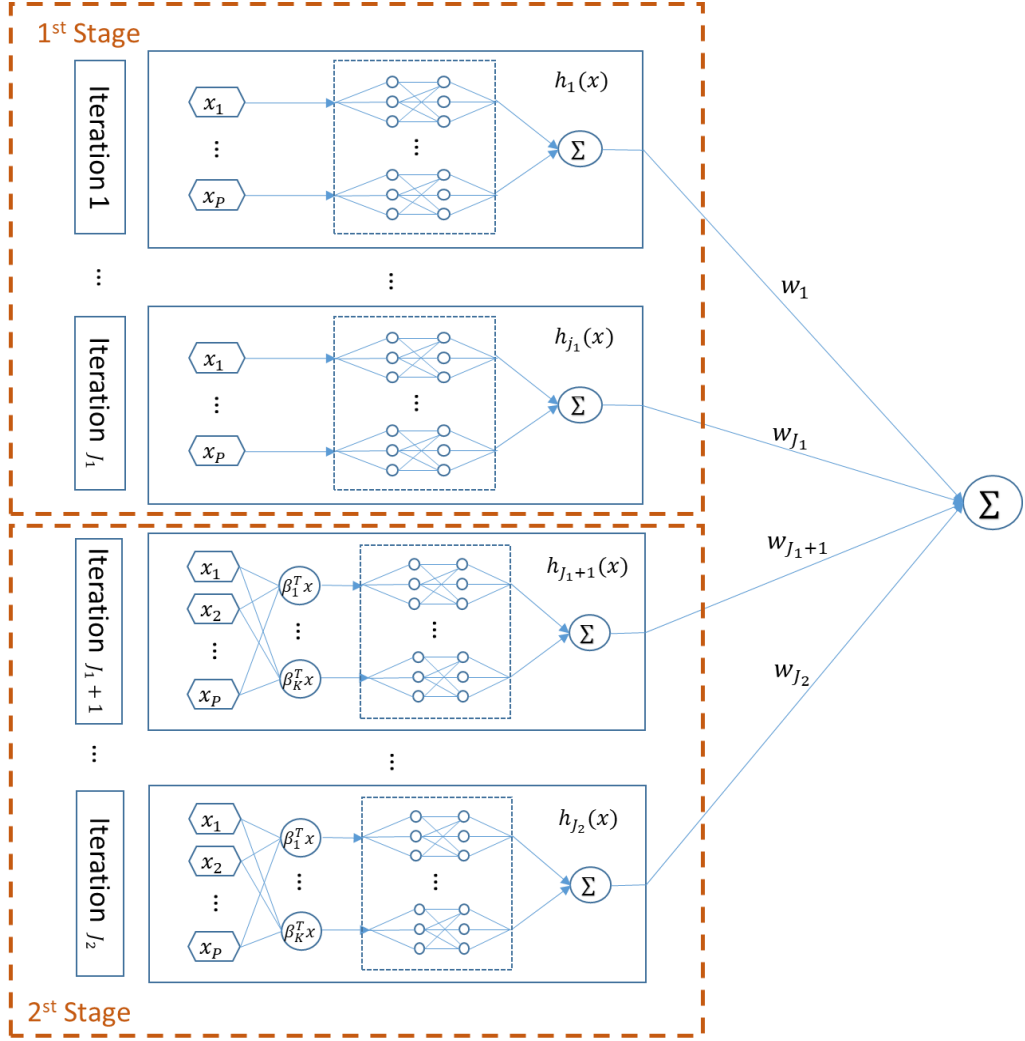

Figure 2: Illustration of AxNN framework

---

**Algorithm 1: AxNN with boosting ensemble[3]**

---

1) For the first stage

   For $k = 1, \ldots, J_1$

   a. Train $h_k(x)$ by $\min_{h_k} \frac{1}{N}\sum_{i=1}^{N} \Phi\left(\sum_{j=1}^{k-1} w_j h_j(x_i) + h_k(x_i), y_i\right)$ with $\sum_{j=1}^{k-1} w_j h_j(x_i)$ fixed, where $h_k(x)$ is GAMnet.

   b. Train $w_1, \ldots w_k$ by $\min_{w_1,\ldots,w_k} \frac{1}{N}\sum_{i=1}^{N} \Phi\left(\sum_{j=1}^{k-1} w_j h_j(x_i) + w_k h_k(x_i), y_i\right)$ with $h_1, \ldots, h_k$ fixed.

2) For the second stage
   Assume $L = \sum_{j=1}^{J_1} w_j h_j(x_i)$ are obtained from the first stage, and fix it.
   For $k = J_1 + 1, \ldots, J_2$

---

[3] All penalty terms are ignored in Algorithms 1 and 2 for simplicity.

a. Train $h_k(x)$ by $\min_{h_k} \frac{1}{N}\sum_{i=1}^{N} \Phi\left(L + \sum_{j=J_1+1}^{k-1} w_j h_j(x_i) + h_k(x_i), y_i\right)$ with $\sum_{j=J_1+1}^{k-1} w_j h_j(x_i)$ fixed, where $h_k(x)$ is xNN.

b. Train $w_{J_1+1}, \ldots w_k$ by $\min_{w_{J_1+1},\ldots w_k} \frac{1}{N}\sum_{i=1}^{N} \Phi\left(L + \sum_{j=J_1+1}^{k-1} w_j h_j(x_i) + w_k h_k(x_i), y_i\right)$ with $h_{J_1+1}, \ldots, h_k$ fixed.

---

### 3.2.2 AxNN with Stacking

The original AdaNet package uses an approach called `stacking' with each base learner trained using the original response variable rather than "residuals" as we did in Section 3.1. For each iteration, AdaNet selects the best subsets among multiple candidate subsets; the candidate subnetworks can vary with different depths and over iterations, usually with increasing complexity manner. Thus, even though AdaNet base learner is trained against the responses, the base learners are different for different iterations. The details with stacking ensemble are given in Algorithm 2.

The rationale here is model (weighted) averaging and stacking, similar to random forest. The base learner from each iteration is unbiased but with high variance, and the variance is reduced through weighted averaging/stacking. This method requires strong base learners: deeper or wider NN architecture. In contrast, the rationale behind boosting is similar to gradient boosting machine (GBM), where start with weak learners and boost performance over the iterations by removing bias through fitting the "residuals".

AxNN with stacking is more sensitive to the initial subnetwork architecture in the first iteration. If the true model is complicated but the first base learner is too weak, the performance can be poor. The weights of the base learners from iterations also behave differently between stacking and boosting. If the base learners get stronger over iterations, their weights with stacking generally decay over the iterations. However, in our studies, the weights of the base learners with boosting are usually stable over the iterations.

---

**Algorithm 2: AxNN with stacking ensemble**

---

1) For the first stage
   For $k = 1, \ldots, J_1$

   a. Train $h_k(x)$ by $\min_{h_k} \frac{1}{N}\sum_{i=1}^{N} \Phi(h_k(x_i), y_i)$, where $h_k(x)$ is GAMnet.

   b. Train $w_1, \ldots w_k$ by $\min_{w_1,\ldots,w_k} \frac{1}{N}\sum_{i=1}^{N} \Phi\left(\sum_{j=1}^{k-1} w_j h_j(x_i) + w_k h_k(x_i), y_i\right)$ with $h_1, \ldots, h_k$ fixed.

2) For the second stage
   Assume $L = \sum_{j=1}^{J_1} w_j h_j(x_i)$ are obtained from the first stage, and fix it.
   For $k = J_1 + 1, \ldots, J_2$

   a. Train $h_k(x)$ by $\min_{h_k} \frac{1}{N}\sum_{i=1}^{N} \Phi(L + h_k(x_i), y_i)$, where $h_k(x)$ is xNN.

b. Train $w_{J_1+1}, \ldots w_k$ by $\min\limits_{w_{J_1+1},\ldots w_k} \frac{1}{N}\sum_{i=1}^{N} \Phi\left(L + \sum_{j=J_1+1}^{k-1} w_j h_j(\boldsymbol{x_i}) + w_k h_k(\boldsymbol{x_i}), y_i\right)$ with $h_{J_1+1}, \ldots, h_k$ fixed.

---

### 3.3 Feature interactions with xNN and AxNN

As noted earlier, in the first stage of AxNN, the GAMnet base-learner captures the overall 'projected' main effects, including the embedded main effects in the interaction terms. In the second stage, AxNN with xNN base learners capture the remaining effects. These effects are estimated via a sum of AIMs using xNN. The details are discussed in Section 3.4. In calculating the interactions, we remove any main effects embedded in the interaction terms. This is important as any embedded main effects may distort the magnitude of the interaction effects. Therefore, most interaction measures in the literature are based on the elimination of main effects first. For example, H-statistics in (Friedman and Popescu 2008) considers the difference between 2-D PDP and the sum of 1D-PDP as interaction strength statistic. When (Tsang, Cheng and Liu 2017) use neural network for interaction detection, they separate the main effects by univariate network, which can reduce creating spurious interactions using the main effects.

### 3.4 Ridge function decomposition for interpretation

Although each base learner of AxNN in the form of GAMnet or xNN is interpretable, there can be multiple iterations in the ensemble, making the interpretation more difficult. When there are $P$ predictors and $K$ ridge functions in a base learner, and $J$ iterations, there will be $K \times J$ ridge functions, and $K \times J \times P$ xNN projection coefficients in total for the second stage, which makes the interpretation more difficult.

To enhance the interpretability of AxNN, we propose decomposing the ridge functions by grouping those with the same projection coefficient patterns. For the first stage, the ridge functions with the same covariate are grouped together to account for the main effect of the corresponding covariate. For the second stage, we apply a coefficient threshold value to the projection layer coefficients of each ridge function, and the projection coefficients bigger than the given threshold value are considered as active, Furthermore, those ridge functions with the same set of active projection coefficients are aggregated. Different sets of active projection coefficients account for different interaction patterns.

More specifically, for the first stage, the main effect of covariate $x_p$, denoted by $M(x_{i,p})$, can be calculated by aggregating all the ridge functions w.r.t $x_p$.

$$M(x_{i,p}) = \sum_{j=1}^{J_1} w_j g_{j,p}(x_{i,p}), \qquad j = 1, \ldots, J_1$$

To calculate the interaction effects, let $S$ denote the set of all the combinations of the predictors. For example, with covariates $\{x_1, x_2, x_3\}$, $S = \{\{x_1\}, \{x_2\}, \{x_3\}, \{x_1, x_2\}, \{x_2, x_3\}, \{x_1, x_3\}, \{x_1, x_2, x_3\}\}$. For any ridge function, we expect only a subset of projection coefficients to be significantly from zero. We will select these using a threshold $\theta > 0$. Define $l(\boldsymbol{\beta}_{j,k})$ as the set of predictors with projection coefficients whose magnitude is greater than $\theta$. Specifically,

$$l(\boldsymbol{\beta}_{j,k}) = \{x_p : |\beta_{j,k,p}| > \theta, p = 1, \ldots, P\}.$$

We call $l(\boldsymbol{\beta}_{j,k})$ the active set of projection indices.

For any interaction term $q \in S$, the corresponding interaction effect is defined as

$$I_q(\boldsymbol{x}_i) = \sum_{j=J_1+1}^{J_2} \sum_{k=1}^{K_j} w_j g_{j,k}(\boldsymbol{\beta}_{j,k}^T \boldsymbol{x}_i) I(l(\boldsymbol{\beta}_{j,k}) = q),$$

where $I(x)$ is one when $x$ is True; and zero otherwise.

Finally, the fitted response $\hat{f}$ can be decomposed into

$$\hat{f} = \sum_{j=1}^{J_1} w_j h_j(\boldsymbol{x}_i) + \sum_{j=J_1+1}^{J_2} w_j h_j(\boldsymbol{x}_i)$$

$$= \sum_{p=1}^{P} M(x_{i,p}) + \sum_{q \in S} I_q(\boldsymbol{x}_i)$$

A major benefit of the ridge function decomposition is that we can visualize each pattern by plotting when the dimension of projection indices is low.

Now, the importance of main and interaction effects can be measured by their standardized variances. Letting $var(\cdot)$ denote the sample variance, we have

$$M_p = \frac{var\left(M(x_{i,p})\right)}{var(\hat{f})}$$

$$I_q = \frac{var\left(I_q(\boldsymbol{x}_i)\right)}{var(\hat{f})}$$

To make the discussion more concrete, we illustrate the ridge function decomposition with a simple example based on three covariates $\{x_1, x_2, x_3\}$. Suppose there are two iterations for the first stage and two iterations for the second stage. For the second stage, we assume there are two ridge functions for xNN. The base learners and ridge function coefficients are listed in Table 1. To make the notation simpler, we remove the subscription $i$ for sample index.

Table 1: An illustrated example of AxNN based on three covariates.

| Iteration ($j$) | Base learner type | Number of ridge functions ($K_j$) | Base learner weight ($w_j$) | $k = 1$ | $k = 2$ | $k = 3$ |
|---|---|---|---|---|---|---|
| 1 | GAMnet | 3 | $w_1$ | $g_{1,1}(x_1)$ | $g_{1,2}(x_2)$ | $g_{1,3}(x_3)$ |
| 2 | GAMnet | 3 | $w_2$ | $g_{2,1}(x_1)$ | $g_{2,2}(x_2)$ | $g_{2,3}(x_3)$ |
| 3 | xNN | 2 | $w_3$ | $g_{3,1}(\boldsymbol{\beta}_{3,1}^T \boldsymbol{x})$, $\boldsymbol{\beta}_{3,1}^T = (0.9, 0.9, 0.01)$ | $g_{3,2}(\boldsymbol{\beta}_{3,2}^T \boldsymbol{x})$, $\boldsymbol{\beta}_{3,2}^T = (0.9, -0.9, 0.05)$ | |
| 4 | xNN | 2 | $w_4$ | $g_{4,1}(\boldsymbol{\beta}_{4,1}^T \boldsymbol{x})$, $\boldsymbol{\beta}_{4,1}^T = (0.08, 0.8, 0.8)$ | $g_{4,2}(\boldsymbol{\beta}_{4,2}^T \boldsymbol{x})$, $\boldsymbol{\beta}_{4,2}^T = (0.01, 0.8, -0.8)$ | |

With threshold $\theta = 0.15$, $\boldsymbol{\beta}_{3,1}$ and $\boldsymbol{\beta}_{3,2}$ have the active set of $\{x_1, x_2\}$, and $\boldsymbol{\beta}_{4,1}$ and $\boldsymbol{\beta}_{4,2}$ have the active set of $\{x_2, x_3\}$. The ridge function decomposition regroups the ridge functions from the base learners from different iterations, and is showed in Table 2. There are three main effects and two interaction terms on $\{x_1, x_2\}$ and $\{x_2, x_3\}$.

Table 2: The ridge function decomposition for the simple AxNN example

| Index | main/interaction effect | Ridge function decomposition |
|---|---|---|
| $p = 1$ | $M(x_1)$ | $w_1 g_{1,1}(x_1) + w_2 g_{2,1}(x_1)$ |
| $p = 2$ | $M(x_2)$ | $w_1 g_{1,2}(x_2) + w_2 g_{2,2}(x_2)$ |
| $p = 3$ | $M(x_3)$ | $w_1 g_{1,3}(x_3) + w_2 g_{2,3}(x_3)$ |
| $q = \{x_1, x_2\}$ | $I_{\{x_1,x_2\}}(\boldsymbol{x})$ | $w_3 g_{3,1}(\boldsymbol{\beta}_{3,1}^T \boldsymbol{x}) + w_3 g_{3,2}(\boldsymbol{\beta}_{3,2}^T \boldsymbol{x})$ |
| $q = \{x_2, x_3\}$ | $I_{\{x_2,x_3\}}(\boldsymbol{x})$ | $w_4 g_{4,1}(\boldsymbol{\beta}_{4,1}^T \boldsymbol{x}) + w_4 g_{4,2}(\boldsymbol{\beta}_{4,2}^T \boldsymbol{x})$ |

As one can see, the threshold value plays an important role in the decomposition results. Too small threshold values can result in many non-zero project coefficients, i.e., many high order interaction effects, while too large threshold values can result in main effects from the second stage, which should be fully captured by the first stage. Our simulation studies revealed that, within a certain range, the ridge decomposition is generally stable. One reason is that the inputs of neural network are typically scaled before training. From our experiments, we recommend a threshold range between 0.15 and 0.3.

As the ridge function decomposition is conducted after training the algorithm and computation is very fast, an empirical approach for determining the threshold value is to try several different candidate threshold values and choose a reasonable value. Alternatively, one can use the histogram of the absolute projection coefficient values to guide our choice of selection values.

### 3.5 Implementation

Our implementation of AxNN is based on modifying Google's open source AutoML tool AdaNet package, version 5.2 (Weill, et al. 2019). AdaNet package is a lightweight TensorFlow-based framework (Abadi, et al. 2016) for automatically learning high-quality models with minimal expert intervention. We modified the source code to accommodate our two-stage approach and both boosting and stacking ensemble approaches. With Google AdaNet implementation, we inherit the benefits of adaptively learning a neural network architecture via multiple subnetwork candidates.

### 3.6 One-stage AxNN

Thus far, we have studied a two-stage AxNN using GAMnet and xNN, but xNN is able to directly capture main effects as well as interactions. Therefore, an alternative is to use a one-stage approach with xNN base learners. For this algorithm, we need to just remove the first stage from Algorithm 1, and set $L$ as 0 in the second stage. Based on our experiments, one-stage AxNN also has reasonable performance.

An advantage of one-stage AxNN is that it does not artificially separate the main effects by projecting into each individual covariate dimension, if the true model form does not have explicit main effect terms. Moreover, one-stage AxNN usually generates more parsimonious representation of main effects and interaction terms in the ridge decomposition in Section 3.4. On the other hand, the interpretation can be more difficult as the main effects and interaction effects may not be easily separable. As mentioned in Section 3.3, the main effects that are embedded in the nonlinear ridge function may contaminate the importance measure of the interaction effects. Further, when the ridge function is linear, multiple main effects can be absorbed into a single ridge function, causing the entanglement of main effects and interaction effects. To make interpretability more stable and clear, we focus on two-stage AxNNs in this paper.

## 4 Experiments

The performance of two-stage AxNN with both boosting stacking are studied on several synthetic examples.

### 4.1 Simple synthetic example

We first use the following simple synthetic example:

$$y = x_1 + x_2^2 + x_3^3 + e^{x_4} + x_1 x_2 + x_3 x_4 + \epsilon,$$

where the predictors are independent and are uniform [-1, 1], the error term is normal $N(0, 0.1)$, and the sample sizes for training, validation, and testing data sets are 50K, 25K and 25K, respectively. We use the boosting case as an illustration.

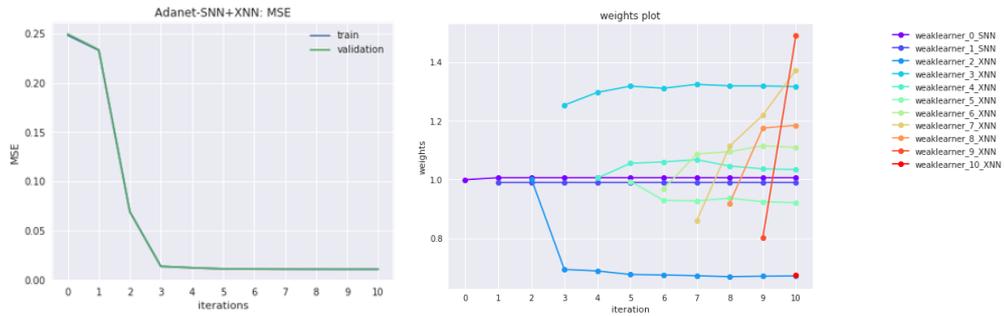

Figure 3: Training and validation loss trajectory (left) and the weights of weak learners over iterations (right). GAMnet is denoted as SNN (structured neural nets) in the plots.

Figure 3 shows there is a steep decrease of training and validation errors after two GAMnet weak learners. We can automatically select the architecture from those of previous iterations and other candidate networks with the same number of layers but with one additional unit. The selected NN types and architectures over the boosting iterations are listed in Table 3.

Table 3: Neural network type and architectures over the iterations

| stage | 1 | 1 | 2 | 2 | 2 | 2 | 2 | 2 | 2 | 2 | 2 | 2 |
|---|---|---|---|---|---|---|---|---|---|---|---|---|
| iteration | 1 | 2 | 1 | 2 | 3 | 4 | 5 | 6 | 7 | 8 | 9 | 10 |
| weak learner type | GAMnet | GAMnet | xNN | xNN | xNN | xNN | xNN | xNN | xNN | xNN | xNN | xNN |
| # of layer | 1 | 1 | 1 | 1 | 1 | 1 | 1 | 1 | 1 | 1 | 1 | 1 |
| # of units | 5 | 6 | 6 | 7 | 8 | 8 | 8 | 9 | 9 | 9 | 10 | 11 |

The mean square error (MSE) and R square(R2) score of the testing data are 0.0107 and 0.9913 respectively, which is very close to the ground truth values of MSE 0.01 and R2 score 0.9919. The importance of the main effects and interaction effects from the ridge function in Figure 3 is well aligned with their true importance, calculated through the variance of each additive component in the true model. We also observe that some active sets of project coefficients from the second stage (e.g.,$(x_1, x_3, x_4)$) make almost no contributions to the response, meaning that the corresponding ridge functions are almost flat and close to 0. In the following figures, for clarity of exposition, insignificant active sets will not be plotted if their contributions are less than 0.1%.

Using the discussion in Section 3.4, the response can be decomposed into main effects and interaction effects in an additive manner. As the order of the interactions is low, all the main and interaction effects can be visualized directly in Figures 4 and 5. The results are consistent with the true model.

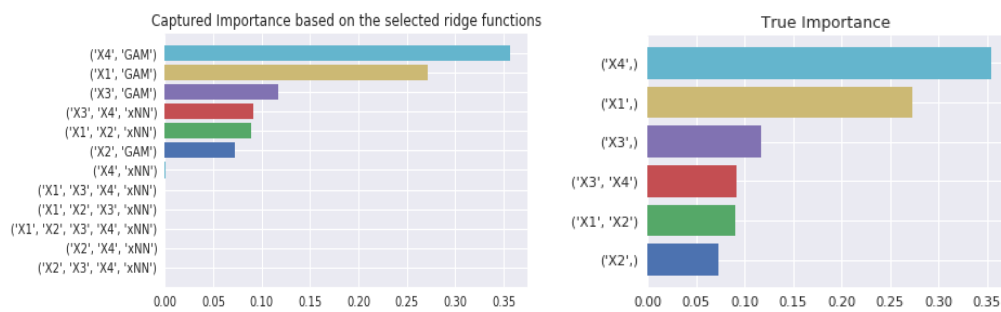

Figure 4: Ridge function decomposition for simple synthetic example

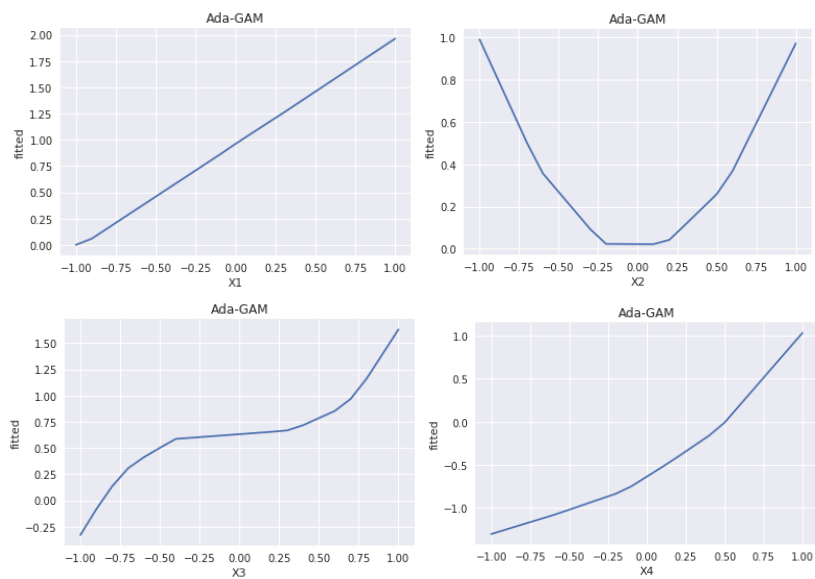

Figure 5: main effects based on the ridge function decomposition

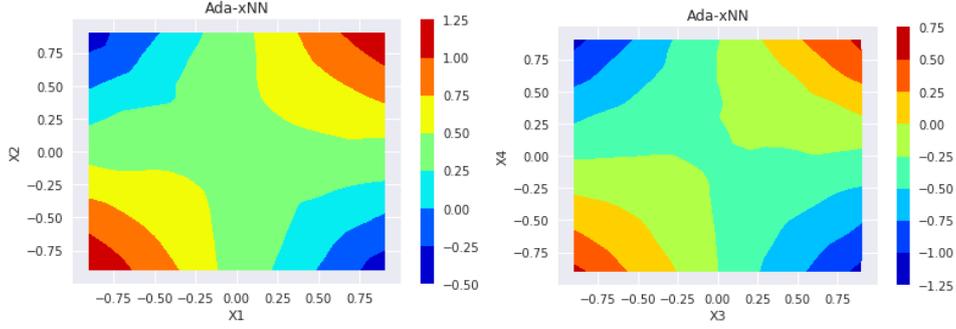

**Figure 6: Interaction effects based on the ridge function decomposition**

## 4.2 More complex examples

We consider four additional examples. The first dataset was generated by a function used in (Hooker 2004), one used widely for interaction detection testing (Sorokina, et al. 2008, Lou, et al. 2013, Tsang, Cheng and Liu 2017). The second case is our own. The third and fourth cases are from (Tsang, Cheng and Liu 2017). In all these, there are ten independent predictors. We used random train/valid/test splits of 50%/25%/25% on 200K data points.

The models are given below:

- **Example 1**

$$f(x) = \pi^{x_1 x_2}\sqrt{2x_3} - \sin^{-1} x_4 + \log(x_3 + x_5) - \frac{x_9}{x_{10}}\sqrt{\frac{x_7}{x_8}} - x_2 x_7,$$

where $x_1, x_2, x_3, x_6, x_7, x_9 \sim U(0,1), x_4, x_5, x_8, x_{10} \sim U(0.6, 1)$.

- **Example 2:**

$$f(x) = x_1^2 + x_2^2 + x_3^2 + x_3 x_4 + 2 x_4 x_5 x_6 + x_4^3 x_7 + x_5 x_6 x_7 + x_7 x_8 x_9 x_{10},$$

where $x_1, \ldots, x_{10} \sim U(-1, 1)$

- **Example 3:**

$$f(x) = x_1 x_2 + 2^{x_3 + x_5 + x_6} + 2^{x_3 + x_4 + x_5 + x_7} + \sin(x_7 \sin(x_8 + x_9)) + \arccos(0.9 x_{10}),$$

where $x_1, \ldots, x_{10} \sim U(-1, 1)$

- **Example 4:**

$$f(x) = \frac{1}{1 + x_1^2 + x_2^2 + x_3^2} + \sqrt{\exp(x_4 + x_5)} + |x_6 + x_7| + x_8 x_9 x_{10},$$

where $x_1, \ldots, x_{10} \sim U(-1, 1)$

For both boosting and stacking ensemble, we considered only one layer for the ridge function subnetworks. The automatic selection of the NN architecture proceeds in the same manner as described in Section 4.1. AxNN boosting starts with weak GAMnet and xNN networks: xNN with 2 subnets and each ridge subnetwork with 3 or 5 units. The stacking AxNN starts with stronger GAMnet and xNN networks: xNN with 15 or 20 subnets and each ridge subnetwork with 10 units.

Results of test performances and comparisons with random forest (RF), XgBoost (XGB) and fully connected feed forward NN (FFNN) are given in Table 4. RF, XgBoost, and FFNN are tuned via grid search. FFNN has two layers with a compatible layer size of AxNN. The AxNN approaches were not tuned extensively as Adanets can do efficient NN architecture search. (say something here about replications and std errors).

AxNN stacking has the best performance over all the four examples. AxNN boosting and FFNN come close. As the true response surfaces are smooth, the tree-based ensemble algorithms do not perform as wells as NN approaches, with RF having the worst performance. The performance of AxNN boosting is not as good as AxNN stacking in these cases, but it is possible they can be improved with further tuning. We do not study this issue further in this paper.

Table 4: Test performance for the complicated synthetic examples (std errors in parenthesis )

| No | metric | ground truth | AxNN boosting | AxNN stacking | RF | XGB | FFNN |
|---|---|---|---|---|---|---|---|
| Example 1 | MSE | 0.0 | 0.0013 | **0.0005** | 0.0112 | 0.002 | 0.0016 |
|  |  | (0.0) | (0.0002) | (0.0) | (0.0002) | (0.0001) | (0.0003) |
|  | R2 score | 1.0 | 0.9985 | **0.9994** | 0.9864 | 0.9976 | 0.9981 |
|  |  | (0.0) | (0.0003) | (0.0) | (0.0002) | (0.0001) | (0.0003) |
| Example 2 | MSE | 0.0 | 0.0044 | **0.0011** | 0.1655 | 0.0115 | 0.0197 |
|  |  | (0.0) | (0.0009) | (0.0002) | (0.0022) | (0.0005) | (0.0048) |
|  | R2 score | 1.0 | 0.993 | **0.9982** | 0.7351 | 0.9815 | 0.9685 |
|  |  | (0.0) | (0.0014) | (0.0004) | (0.0041) | (0.0008) | (0.0076) |
| Example 3 | MSE | 0.0 | 0.0028 | **0.0007** | 0.1887 | 0.0153 | 0.0098 |
|  |  | (0.0) | (0.0005) | (0.0002) | (0.0018) | (0.0006) | (0.0018) |
|  | R2 score | 1.0 | 0.9993 | **0.9998** | 0.9537 | 0.9963 | 0.9976 |
|  |  | (0.0) | (0.0001) | (0.0) | (0.0005) | (0.0001) | (0.0004) |
| Example 4 | MSE | 0.0 | 0.002 | **0.0009** | 0.0513 | 0.0036 | 0.0027 |
|  |  | (0.0) | (0.0007) | (0.0002) | (0.0002) | (0.0002) | (0.0003) |
|  | R2 score | 1.0 | 0.9958 | **0.9981** | 0.8935 | 0.9926 | 0.9944 |
|  |  | (0.0) | (0.0014) | (0.0005) | (0.0005) | (0.0004) | (0.0005) |

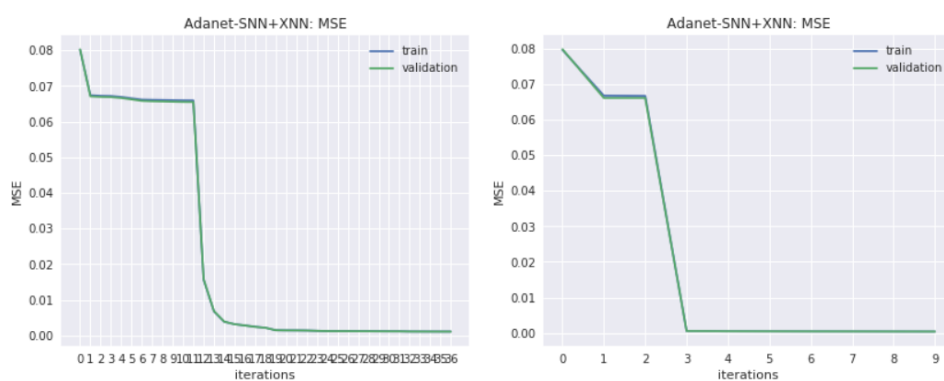

Figure 7: Training and validation loss over two stages for the first synthetic example
—Boosting ensemble (left) and stacking ensemble (right)

Figure 7 shows that AxNN boosting and stacking have similarities as well as differences in their behavior over iterations. Both have similar convergence behavior in each of the two stages and exhibit a steep decrease of validation error at the beginning of the second stage. But boosting converges much slower. This is likely due to the use of weak learners and the boosting mechanism.

To study the behavior of AxNN ridge functions, we first decompose the interaction term of the true model form into projected main effect and the remaining interaction effects, also called pure interaction effects. As it is not straightforward to obtain the analytical form of the projected main effect of each interaction term, we use GAMnet to get it numerically. Then, we aggregate all the projected main effects and original main effect of the same variable and generate the overall projected main effects. This allows us to compare the effects from the AxNN ridge decomposition with the projected main effects and pure interaction effects from the true model form in two approaches.

First, we generate the similar importance plot for the decomposed effects from the true model form, and compare to the importance from AxNN ridge function decomposition. Second, to further evaluate the relationship between AxNN main/interaction effects and the true ones, for each AxNN main or interaction effect, we calculate the correlations with all the effects from the true model, and find and list the true model effect with the maximum correlation.

The ridge function decomposition results for the four synthetic examples are shown in Figure 8, Figure 9, Figure 10, and Figure 11 respectively. The left two plots in each figure show the importance measures from the AxNN boosting and stacking ensemble, respectively. The effects next to the left y-axis are the ranked main or interaction effects from the ridge function decomposition. The labels next to the right y-axis list the corresponding true effects with the maximum correlations. The rightmost plot depicts the true importance. Both boosting and stacking give reasonable main effect and interaction effect estimation from the ridge function decomposition. For all the four synthetic examples, almost all the main effects from the first stage correctly capture the true projected main effects (correlation close to 1).

The second stage is also able to detect and capture the significant high-order interactions correctly (high correlations with the true pure interaction terms for all the four synthetic examples). The estimation of the insignificant interactions are less accurate and unstable. In the first synthetic example, the top interactions $(x_1, x_2, x_3)$ and $(x_7, x_8, x_9, x_{10})$ are correctly detected, and the estimated pure interaction effects have strong correlations over 0.8 with the true ones. However, the weak interaction effects $(x_2, x_7)$ and $(x_3, x_5)$ are missed. Furthermore, when the interactions have a big overlap, the union of the interactions (with higher order) can be detected instead. For example, in the second synthetic example, true interaction effect $(x_4, x_5, x_6)$ and $(x_5, x_6, x_7)$ are captured by their union $(x_4, x_5, x_6, x_7)$ in the ridge function decomposition, and the true effect $(x_4, x_5, x_6)$ is listed due to its importance. However, if a large project threshold is applied, $(x_4, x_5, x_6)$ instead of $(x_4, x_5, x_6, x_7)$ will be shown as the first.

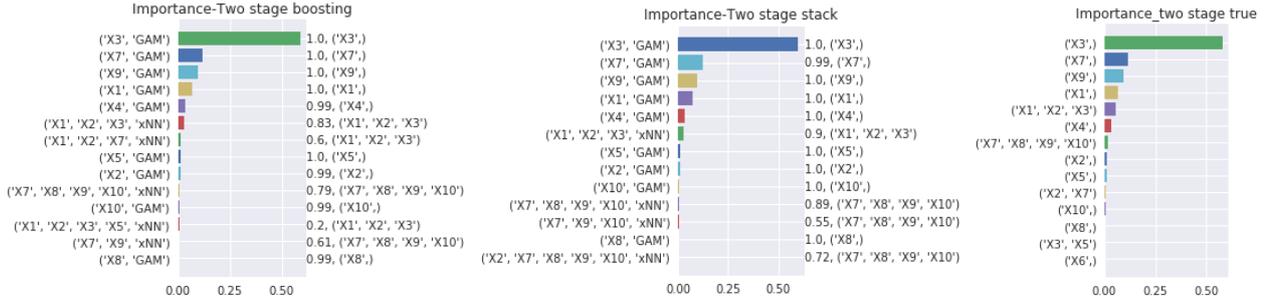

Figure 8: Ridge function decomposition for Synthetic Example 1

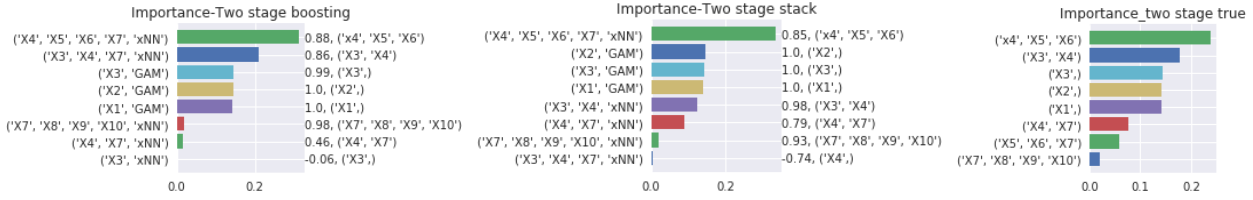

Figure 9: Ridge function decomposition for Synthetic Example 2

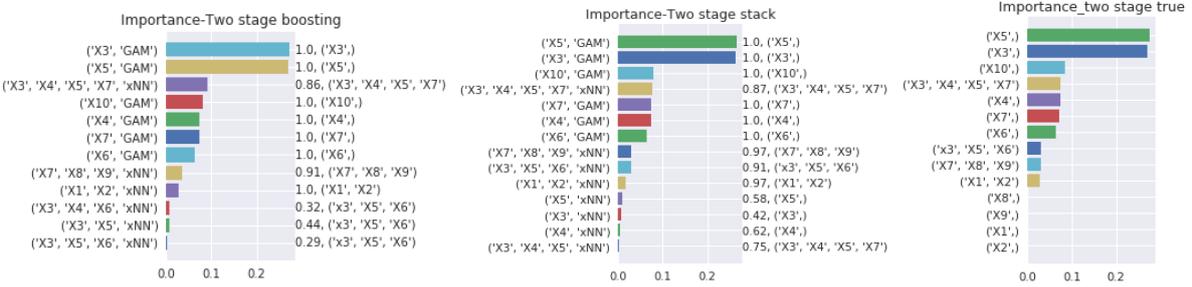

Figure 10: Ridge function decomposition for Synthetic Example 3

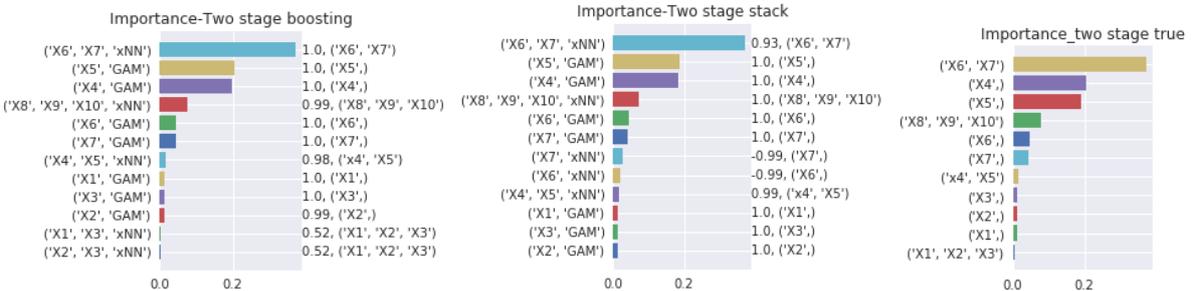

Figure 11: Ridge function decomposition for Synthetic Example 4

To evaluate AxNN performance in the presence of noise, we add normally distributed errors with standard deviation 0.5 to all the four synthetic examples and re-test the performances. Table 5 shows that the performances are consistent with those for the non-error cases in Table 4. Ridge function decomposition result is also generally consistent with the true model, but sometimes a little weaker in the presence of noise. These results are not shown here.

Table 5: Test performance for the complicated synthetic examples ( std errors in paranthesis)

| No | metric | ground truth | AxNN boosting | AxNN stacking | RF | XGB | FFNN |
|---|---|---|---|---|---|---|---|
| Example 1 | MSE | 0.251 (0.002) | 0.256 (0.002) | **0.254** (0.002) | 0.265 (0.002) | 0.256 (0.002) | **0.254** (0.002) |
| | R2 score | 0.766 (0.002) | 0.762 (0.002) | **0.764** (0.002) | 0.753 (0.002) | 0.762 (0.002) | **0.764** (0.002) |
| Example 2 | MSE | 0.251 (0.002) | 0.263 (0.002) | **0.257** (0.002) | 0.413 (0.003) | 0.27 (0.002) | 0.274 (0.004) |
| | R2 score | 0.713 (0.003) | 0.7 (0.003) | **0.706** (0.003) | 0.528 (0.005) | 0.691 (0.003) | 0.687 (0.004) |
| Example 3 | MSE | 0.251 (0.002) | 0.258 (0.002) | **0.255** (0.002) | 0.436 (0.003) | 0.269 (0.002) | 0.264 (0.003) |
| | R2 score | 0.942 (0.001) | **0.941** (0.001) | **0.941** (0.001) | 0.899 (0.001) | 0.938 (0.001) | 0.939 (0.001) |
| Example 4 | MSE | 0.251 (0.002) | 0.256 (0.002) | **0.254** (0.002) | 0.306 (0.002) | 0.259 (0.002) | 0.257 (0.002) |
| | R2 score | 0.658 (0.002) | 0.651 (0.002) | **0.653** (0.002) | 0.583 (0.002) | 0.646 (0.002) | 0.649 (0.003) |

## 5  Applications

We illustrate the results on three applications on binary regression (leading to classification). Two of them have been previously discussed in the literature: i) bike sharing data (Fanaee-T and Gama 2014) and ii) Higgs-Boson data (Adam-Bourdarios 2014). The third one is an application to home mortgages. For all the three datasets, we used random splits into train/valid/test sets of 50%/25%/25%. The starting GAMnet and xNN network architecture is similar to the synthetic examples in Section 4.2.

Table 6: Performance on test data for bike share, Higgs Boson and mortgage examples

| | N | p | Metrics | AxNN boosting | AxNN stacking | RF | XGB | FFNN |
|---|---|---|---|---|---|---|---|---|
| Bike share | 17,379 | 11 | MSE | 0.084 (0.005) | 0.075 (0.003) | **0.074** (0.003) | 0.077 (0.004) | 0.133 (0.005) |
| | | | R2 score | 0.916 (0.003) | **0.925** (0.002) | **0.925** (0.001) | 0.923 (0.002) | 0.867 (0.003) |
| Mortgage | 1,000,000 | 14 | AUC | 0.847 (0.0038) | 0.8469 (0.0033) | 0.8468 (0.0028) | **0.8522** (0.0038) | 0.8461 (0.0045) |
| | | | Logloss | 0.0466 (0.0008) | 0.0469 (0.0008) | 0.0466 (0.0008) | **0.0463** (0.0009) | 0.047 (0.001) |
| Higgs Boson | 818,238 | 30 | AUC | 0.902 (0.002) | 0.905 (0.001) | 0.909 (0.0) | **0.913** (0.0) | 0.91 (0.001) |
| | | | Logloss | 0.372 (0.004) | 0.365 (0.002) | 0.359 (0.001) | **0.35** (0.001) | 0.357 (0.001) |

## 5.1 Bike sharing data

The bike sharing dataset contains 17k data points, and the goal is to predict the hourly count of rental bikes in different weather environments. We used log-counts as the response to reduce skewness. We removed some non-meaningful information as well as two response-related information, which left us with 11 predictors to model. Table 6 shows the performance on the test dataset for the bike data. RF and AxNN stacking have the best predictive performance, followed by Xgboost and AxNN boosting, and FFNN has the worst performance.

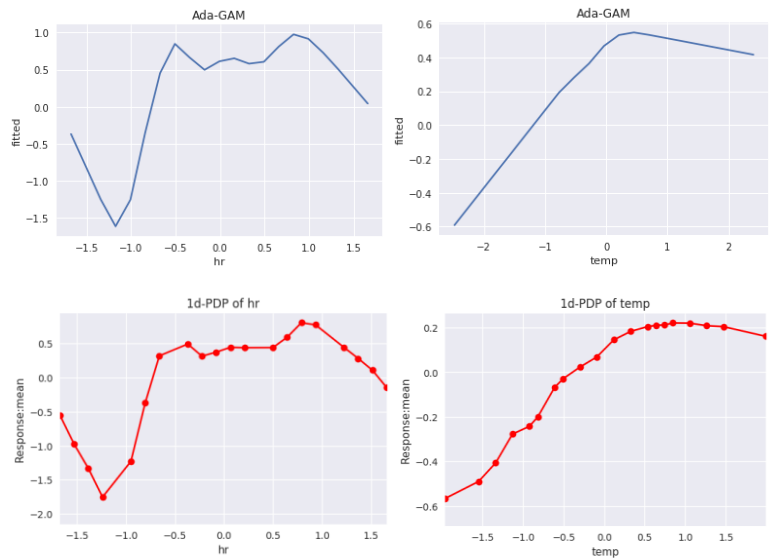

Figure 12: Top two projected main effects for bike share data: AxNN boosting (top) and PDP from RF (bottom)

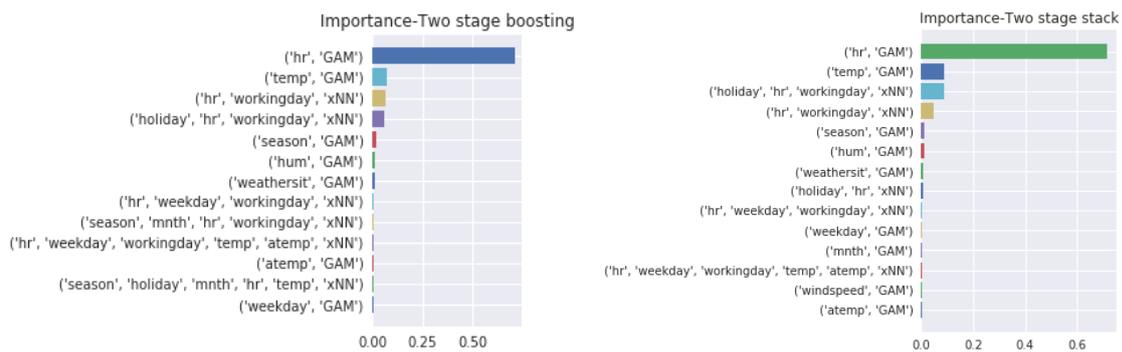

Figure 13: Ridge function decomposition for bike share data: AxNN boosting (left) and AxNN stacking (right)

Figure 14: H-statistics of random forest for bike share data: original scaled (left) and unscaled (right)

Figure 12 depicts the top two projected main effects of 'hr' and 'temp' for bike sharing data, and they are consistent with the 1-D PDP plots for random forest. However, we had to remove two variables-- 'mnth' and 'atemp' which are highly correlated with 'season' and 'temp'-- from the RF training to avoid problems with extrapolation in constructing the PDPs (see (Apley 2016, Liu, et al. 2018)). This reduced RF's testing R2 value to 0.924. The results for AxNN are illustrated in Figure 13. Both AxNN approaches detect the following main effects and have similar rankings: {'hr', 'temp', 'season', 'hum', 'weathersit', 'atemp' and 'weekdays'}. The top interaction effects identified by both are ('holiday', 'hr', 'working day') and ('hr', 'working day'). There are a few smaller ones including the five-factor interactions ('hr,' 'weekday', 'workingday', 'temp', 'atemp'). To compare these results, we calculated the H-statistics (both scaled and unscaled). The top pairs of H-statistics are ('hr', 'working day'), ('working day', 'holiday') ('weekday', 'working day') and these are consistent with the two-factor interactions identified by AxNN. But our approach is able to identify higher-order interactions easily, while it is computationally expensive to calculate high-order H-statistics.

## 5.2 Mortgage data

The second dataset is from the business line of home lending for residential mortgage. For illustration purposes, we used a randomly selected subset of one million observations from one portfolio segment. There are 14 predictors and some key ones are explained in Table 7. The goal is to predict the probability of default for the loans over the next nine quarter prediction horizons based on various loan characteristics (e.g., fico, loan-to-value ratio, etc.) as well as macro-economic variables (e.g., unemployment rate).

Table 7: Variable definition for mortgage data

| Variable | Definition |
| --- | --- |
| fico0 | FICO at prediction time |
| ltv_fcast | Forecasted ratio of loan to value (ltv) |
| dlq_new_delq0 | Indicator of whether loan is delinquent (=0) or not (=1) at prediction time |
| unemprt | unemployment rate |
| grossbal0 | gross loan balance |
| h | Prediction horizon |
| premod_ind | Indicator: 1 if before 2007Q2 (financial crisis); 0 if after |

The performance of test dataset in Table 6 shows that XgBoost is the best for mortgage data. But the performances of AxNN boosting and stacking are very competitive and pretty close to RF and FFNN. They have the advantage of being interpretable.

The results of the ridge-function decomposition for the mortgage dataset are shown in Figure 15. Just like the previous examples, both AxNN approaches give a consistent ordering of the main effects. Moreover, both algorithms identify the top interactions (dlq_new_delq0, h), (fico0, ltv_fcast) and (fico0, sato2). The main effects of the top three variables-- ltv_forcast, FICO0 and dlq_new_delq0-- from AxNN boosting are plotted in Figure 16. The increasing trend of fitted probability of default over ltv_forcast implies that the higher loan to value ratio is, the higher default risk is; while the decreasing trend on FICO0 indicates the higher default risk for the loans with lower credit scores. Moreover, being delinquent at prediction time can also a potential indicator of default in future. The pure interaction effect of fico0 and ltv_fcast from AxNN boosting is plotted via the contour plot in Figure 17, where the positive slopes on ltv_fcast for high FICO (e.g. at 800) indicated by the color changing from blue to red implies that the loans with high FICO are more sensitive to the change of loan to value ratio. The decreasing trend of fitted probability of default for delinquent loans (dlq_new_delq0=0) over prediction horizons (h) in Figure 17 implies that bad loans will terminate and the quality of loans will improve in the future; while the increasing trend for current loans (dlq_new_delq0=1) shows the deterioration of the quality of current loans over the time as loans will start to be delinquent. Note that the surface for the interaction effects can be volatile as the mortgage data is extremely imbalanced with 0.1% default records.

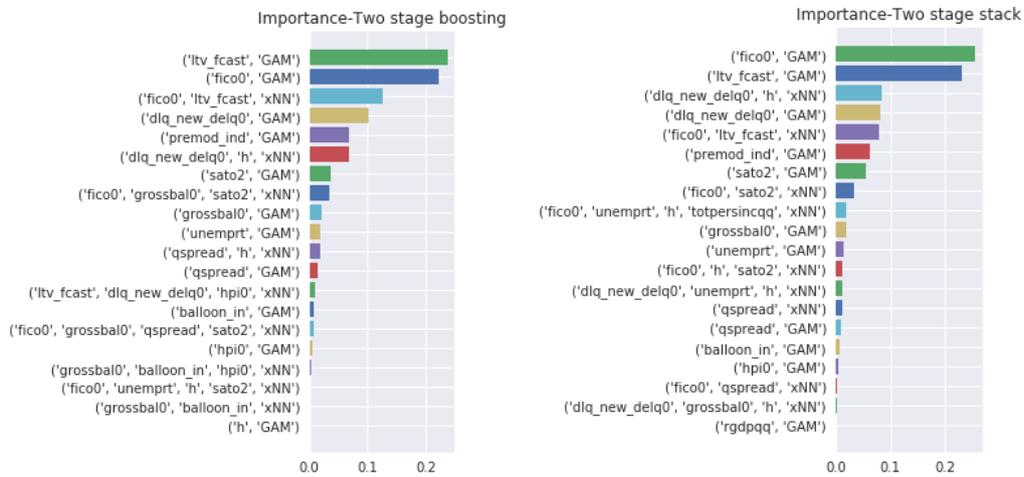

Figure 15: Ridge function decomposition for Mortgage data

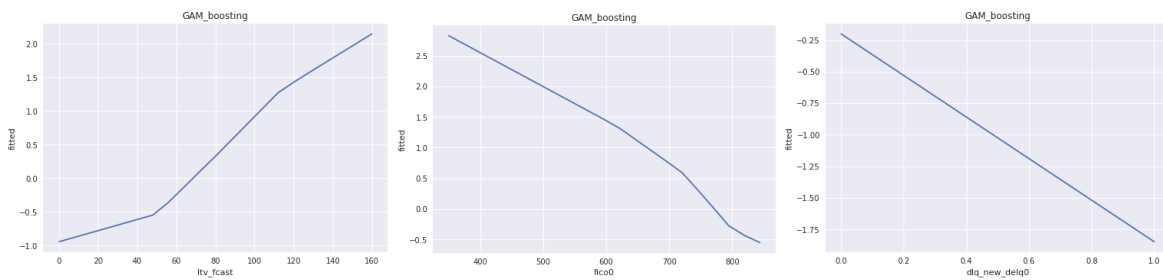

Figure 16: main effects based on the ridge function decomposition for mortgage data

(Left to right : ltv_cast, fico0, dql_new_delq0)

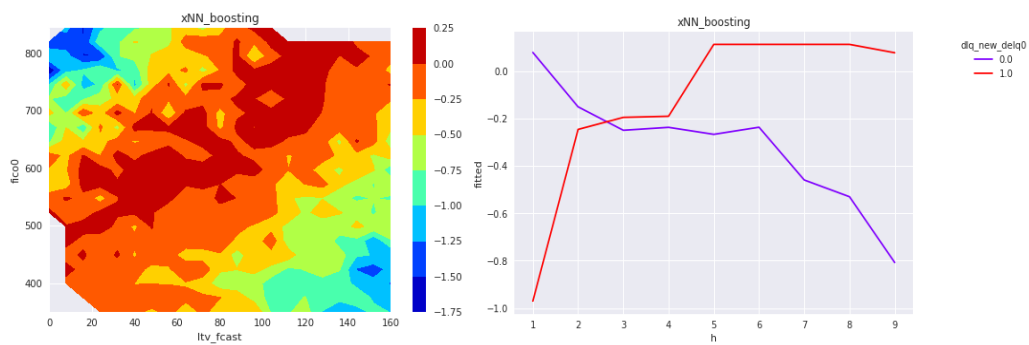

Figure 17: interaction effects based on the ridge function decomposition for mortgage data

(Left: contour plot for fico0 vs ltv_fcast; Right: fitted probability of default vs h with legend on dql_new_delq0)

## 5.3 Higgs_Boson data

This Higgs-Boson dataset has 800k data points and 30 predictors. The goal was to classify the observed events as a signal (new event of interest) or background (something produced by already known processes). Table 6 shows XgBoost the performance on the test dataset for Higgs-Boson data, but the performances of AxNN boosting and stacking are close and competitive.

Some of the predictors in the Higgs-Boson data are highly correlated. Figure 18 shows the identified main effects and interactions. Both AxNN approaches give a consistent ordering of the main effects. The detection of interaction is challenging in the presence of highly correlated predictors. Nevertheless, we detect similar top interaction effects from both: (DER_mass_vis, DER_mass_transverse_met_lep), (DER_mass_vis, DER_deltar_tau_lep, PRI_met), (DER_mass_vis, DER_deltar_tau_lep, DER_pt_ratio_lep_tau). The variables DER_mass_vis and DER_deltar_tau_lep have strong correlation, so their effects may be entangled.

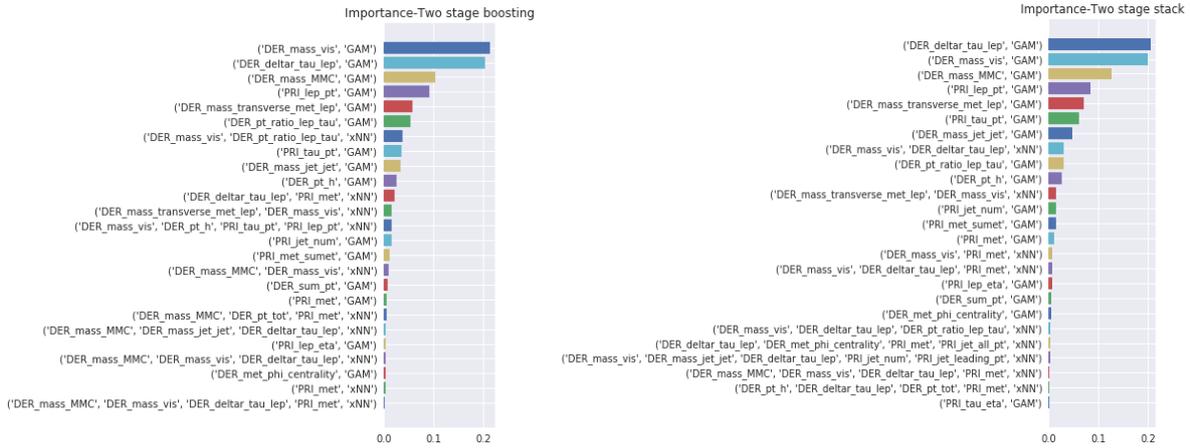

Figure 18: Ridge function decomposition for Biggs Bosons data (top25)

## 6 Concluding Remarks

AxNN is a new machine learning framework that achieves the dual goals of predictive performance and model interpretability. We have introduced and studied the properties of two-stage approaches, with GAMnet base learners to capture the main effects and xNN base learners to capture the interactions. The stacking and boosting algorithms have comparable performances. Both decompose the fitted responses into main effects and higher-order interaction effects through ridge function decomposition. AxNN borrows strength of AdaNet and does efficient NN architecture search and requires less tuning.

## 7 Acknowledgements
We are grateful to Zhishi Wang who contributed to this research while he was at Wells Fargo and to Ming Yuan for useful suggestions.